\pdfoutput=1

\documentclass[11pt]{article}

\usepackage[final]{acl}

\usepackage{times}
\usepackage{latexsym}

\usepackage[T1]{fontenc}

\usepackage[utf8]{inputenc}

\usepackage{microtype}

\usepackage{inconsolata}

\usepackage{graphicx}

\usepackage{listings}
\usepackage{xcolor}
\usepackage{tcolorbox}

\tcbuselibrary{listings,skins}

\usepackage{listings}
\usepackage{xcolor}

\usepackage{xurl}

\lstdefinestyle{framedPython}{
    backgroundcolor=\color{gray!10},   
    frame=single,                      
    rulecolor=\color{gray!70},         
    basicstyle=\ttfamily\small,
    keywordstyle=\color{blue},
    commentstyle=\color{gray},
    stringstyle=\color{red},
    numberstyle=\tiny\color{gray},
    stepnumber=1,
    numbersep=5pt,                     
    breaklines=true,
    showstringspaces=false
}

\title{Trallie: from Free-Form Text to Structured Data with LLMs}

\author{
 \textbf{Vijayasri Iyer\textsuperscript{1}},
 \textbf{Cristiano De Nobili\textsuperscript{1,3}},
 \textbf{Mubashir Hussain Shah\textsuperscript{1}},
 \textbf{Sébastien Bratières\textsuperscript{1,2}}
\\
 \textsuperscript{1}Pi School,
 \textsuperscript{2}Translated,
 \textsuperscript{3}Critiqality
\\
}

\begin{document}
\maketitle
\begin{abstract}
\textsc{Trallie} is an open-source framework that leverages large language models (LLMs) to perform information extraction (IE) from natural language sources with or without user guidance. Unlike conventional approaches that rely on extensive labeled datasets or domain-specific training, \textsc{Trallie} enables the conversion of unstructured data into structured formats with minimal or no representative examples. The framework is designed with three key objectives: (i) reducing manual annotation by harnessing prompt-based zero-shot and few-shot capabilities of modern LLMs, (ii) flexibly parsing diverse document formats (i.e., HTML, PDF, TXT, and JSON), and (iii) improving accessibility for non-expert users by automatically inducing schemas from document collections, which can then be refined for extraction. We evaluate \textsc{Trallie} with open-weight LLMs on FDA 510(k) documents and on movie and university websites from SWDE. It outperforms previously reported systems when the target attributes are provided for text extraction and is competitive when it must discover attributes for the schema automatically. The project is available at \url{https://github.com/PiSchool/trallie}.
\end{abstract}


\section{Introduction}
A substantial portion of valuable information in domains such as healthcare, law, finance, and scientific publishing remains locked inside unstructured or semi-structured documents. Such documents resist efficient searching, querying, and automated analysis, creating barriers to knowledge discovery and decision-making. Transforming this content into structured, queryable formats is fundamental to advancing data-driven applications.

Traditional information extraction (IE) approaches face two major limitations. First, they rely on extensive manual schema design and heavily engineered pipelines, making them costly to develop and difficult to adapt to new domains. Second, they depend on fixed annotation schemas or require costly re-annotation and retraining when custom schemas are needed, offering limited transfer learning benefits even with state-of-the-art models. These limitations are especially evident in legacy natural language descriptions of resources, where structured information remains embedded in narrative text. Existing LLM-based tools also require expertise to use where users must write prompts, design target schemas or provide annotated examples for each new task. Many are additionally closed-source or restricted to proprietary models.

To address these challenges, we introduce \textsc{Trallie}, a lightweight and modular open-source framework that leverages large language models to automatically induce schemas from document collections and populate them with extracted attribute values. Our approach simplifies information extraction by requiring minimal manual annotation while providing flexibility across document formats and accessibility to non-expert users. \textsc{Trallie} operates on three core principles: (1) minimal manual annotation through few-shot learning with well-trained LLMs, (2) format and language flexibility across HTML, PDF, TXT, and JSON, and (3) accessibility to non-experts by automatically inferring schemas from document collections rather than requiring domain expertise.

Our primary contributions are as follows. First, we present a unified 
framework for joint schema generation and extraction from unstructured 
document collections, reducing the need for manual schema design. 
Second, we introduce a provider-agnostic architecture that enables switching between commercial LLM APIs (e.g., OpenAI and Groq), custom endpoints (e.g., Hugging Face), and local models (e.g., Ollama) without code changes, alongside long-document handling through overlapping chunking with LLM-based consolidation, JSON output validation, and multi-format document processing. Third, we provide a complete open-source implementation including a Python package, evaluation scripts, and an interactive Streamlit interface for schema discovery and data extraction.

\section{\textsc{Trallie}}
\textsc{Trallie} operates through a two-stage pipeline: schema generation followed by data extraction. The \textbf{Schema Generator} identifies relevant attributes from a collection of unstructured documents, producing the schema. The \textbf{Data Extractor} then uses this schema to retrieve attribute values from all documents and output structured JSON objects. By decoupling these stages, \textsc{Trallie} enables flexible adoption of either or both components, making it easy to adapt the system to new domains with minimal manual effort.

\begin{figure*}[t]
    \centering
    \includegraphics[width=\linewidth]{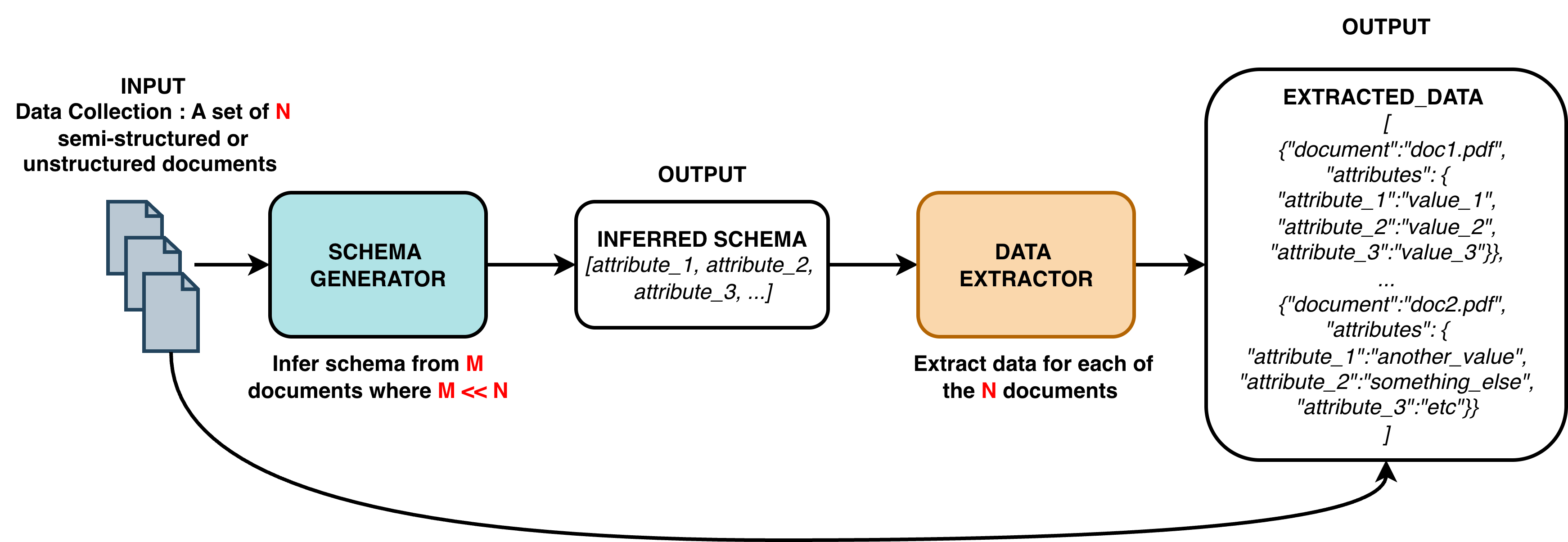}
    \caption{Two-stage pipeline for schema inference and data extraction.}
    \label{fig:architecture}
\end{figure*}

\subsection{Schema Generation}

The Schema Generator identifies attributes that capture the structure 
of a document collection through a three-step process: sampling, 
attribute extraction, and aggregation. The code below shows the simple flow for Schema Generation using the \textsc{Trallie} framework.

\begin{lstlisting}[style=framedPython, language=Python]
from trallie import SchemaGenerator

records = ["data/test_data_1.txt",
           "data/test_data_2.pdf"]

schema_generator = SchemaGenerator(
    provider="openai", 
    model_name="gpt-4o")

schema = schema_generator.discover_schema(
    description="FDA 510(k) clearance documents",
    records=records)
\end{lstlisting}

\paragraph{Step 1: Sampling.}
To reduce computational cost, schema discovery is run on a small subset of the collection rather than on every document. By default, the first 10 documents are used.

\paragraph{Step 2: Attribute Extraction.}
For each sampled document, the LLM proposes a candidate set of attributes that capture the document's information structure (e.g., ``title'', ``authors'', ``abstract'', ``methodology'', and ``conclusions'' for a scientific paper).

\paragraph{Step 3: Schema Aggregation.}
The system aggregates candidate attributes across all sampled documents and selects the top-k most frequently occurring attributes (k = 20 by default). This produces a compact schema that reflects the dominant structure of the collection.

\subsection{Data Extraction}
The Data Extractor takes a document collection and a schema (either 
generated by the Schema Generator or provided by the user) and produces 
structured JSON outputs, one JSON object per document. The code below shows the simple flow for Data Extraction using the \textsc{Trallie} framework. 

\begin{lstlisting}[style=framedPython, language=Python] 
from trallie import DataExtractor 

records = [ "data/test_data_1.txt", 
            "data/test_data_2.pdf", ] 
            
schema = ["title", "authors", "date"] 

data_extractor = DataExtractor(
                provider="openai", 
                model_name="gpt-4o") 
                
for record in records: 
    extracted_json = data_extractor.extract_data(schema, record) 

\end{lstlisting}

\paragraph{Step 1: Document Parsing.}
The system iterates through each document and converts it to plain text according to its format (i.e., PDF, HTML, TXT, or JSON).

\paragraph{Step 2: Attribute Value Extraction.}
The LLM receives the document text together with the schema and returns a JSON object containing a value for each attribute (e.g., given a schema with the attributes ``title'', ``authors'', and ``publication\_date'', the system locates these fields in the document and extracts their values). Values are copied verbatim from the document, and attributes not present in the document are returned as null. The response is parsed as JSON, and the request is retried when the output is not valid JSON.

\subsection{Data Handling}
\textsc{Trallie} supports multiple document formats, including HTML, PDF, TXT, and JSON. For binary formats such as PDF, \textsc{Trallie} has a built-in parser based on PyPDF2 for text extraction. 

\subsection{Long Document Handling}

For documents that exceed a configurable length, \textsc{Trallie} splits the text into overlapping chunks (100k characters with 10k overlap by default, breaking at sentence ends where possible), extracts from each chunk, and consolidates the per-chunk results with a final LLM call.

\subsection{Multiple LLM Backend Support}
A key design principle of \textsc{Trallie} is LLM provider independence. The 
framework implements a unified LLM interface that abstracts over 
different providers, enabling users to switch between commercial 
services (OpenAI, Groq), custom deployments (Hugging Face Inference 
Endpoints), and local models (Ollama) through a single configuration 
parameter. This design choice addresses practical deployment scenarios 
where users may face different constraints: API cost limits, data 
privacy requirements, internet connectivity, or institutional policies.

Additionally, \textsc{Trallie} parses every model response as JSON and retries the request when the output is malformed, so downstream applications receive well-formed JSON regardless of the underlying provider.

\subsection{Implementation and Accessibility}
\textsc{Trallie} is released as an open-source Python package under the Apache 2.0 license. The implementation provides multiple access points: a programmatic API for pipeline integration, a lightweight Streamlit web interface for browser-based interaction, example Jupyter notebooks for common workflows, and reproducible evaluation scripts. 



\section{Evaluation}
We evaluate \textsc{Trallie} on two complementary information-extraction settings: OpenIE and ClosedIE. These tasks jointly assess the system’s ability to both discover schemas from unstructured documents and extract structured attribute values when a schema is provided. In the ClosedIE setting, the target schema is predefined and supplied to the model, which must restrict itself to extracting values only for these attributes without introducing new fields or paraphrases. In contrast, the OpenIE setting provides no schema, requiring \textsc{Trallie} to infer an appropriate set of attributes and populate them directly from the documents.

\subsection{Evaluation Methods}
We use the metrics and released evaluation code of \citet{Arora2023LanguageME}. ClosedIE is scored with token-level Text F1 \citep{rajpurkar2016squad}, averaged over documents for each attribute and then over attributes. OpenIE is scored with Pair F1 over (document, attribute, value) triples, taking the top-$k$ discovered attributes, with $k$ the number of gold attributes.

\subsection{Evaluation Datasets}

We evaluate on the EVAPORATE benchmark \citep{Arora2023LanguageME}: FDA 510(k) and 13 websites from the movie and university verticals of SWDE \citep{hao2011swde}, 1{,}400 documents in total. The datasets are available at \url{https://huggingface.co/datasets/hazyresearch/evaporate}.

\paragraph{Regulatory Domain.}
FDA 510(k) medical device clearance documents; all 100 documents are used.

\paragraph{SWDE Movie Domain.}
AllMovie, AMC TV, Hollywood, iHeartMovies, IMDb, Metacritic, Rotten Tomatoes, and Yahoo Movies. Each website has roughly 2{,}000 pages; we use the first 100 by file name.

\paragraph{SWDE University Domain.}
CollegeProwler, eCampusTours, Embark, MatchCollege, and US News, with 1{,}000--2{,}000 pages each; we use the first 100 by file name.

\section{Results}
\label{sec:results}

\begin{table*}[t]
  \centering
  \small
  \setlength{\tabcolsep}{5pt}
  \begin{tabular}{l l c c c c c c}
    \hline
     & & \multicolumn{2}{c}{\textbf{FDA 510(k)}} & \multicolumn{2}{c}{\textbf{SWDE Movie}} & \multicolumn{2}{c}{\textbf{SWDE University}} \\
    \textbf{System} & \textbf{LLM} & \textbf{ClosedIE} & \textbf{OpenIE} & \textbf{ClosedIE} & \textbf{OpenIE} & \textbf{ClosedIE} & \textbf{OpenIE} \\
    \hline
    DOM-LM$^{\dagger}$ & -- & -- & -- & 71.9 & 54.1 & 68.0 & 55.2 \\
    EVAPORATE-Direct$^{\dagger}$ & text-davinci-003 & 78.0$^{a}$ & 45.5 & 84.4 & 45.2 & 72.6 & 53.8 \\
    EVAPORATE-Code+$^{\dagger}$ & text-davinci-003 & 80.1 & 62.8 & 79.5 & \textbf{56.8} & 73.7 & 59.0 \\
    \hline
    \textsc{Trallie} (ours) & gpt-oss-120b & 95.9 & 68.6 & 93.2$^{b}$ & 49.8$^{b}$ & 89.9$^{b}$ & 56.3$^{b}$ \\
    \textsc{Trallie} (ours) & Qwen3.8-27B & \textbf{97.2} & \textbf{80.2} & \textbf{93.4}$^{b}$ & 52.7$^{b}$ & \textbf{91.5}$^{b}$ & \textbf{64.4}$^{b}$ \\
    \hline
  \end{tabular}
  \caption{ClosedIE (Text F1) and OpenIE (Pair F1) on the EVAPORATE benchmark. SWDE scores are the mean over websites (8 movie, 5 university).
    $^{\dagger}$Reported by \citet{Arora2023LanguageME}; not re-run.
    $^{a}$Per-attribute direct prompting (their Table 6).
    $^{b}$First 100 pages per website (EVAPORATE-Code+ is evaluated on all pages); 95\% bootstrap CIs over pages are within $\pm$0.4 of each mean.
    \textsc{Trallie}'s prompts were developed on FDA 510(k); SWDE was not used during development.
    Systems use different LLMs, so differences reflect both method and model.}
  \label{tab:main-results}
\end{table*}

\begin{table}[t]
  \centering
  \small
  \setlength{\tabcolsep}{4pt}
  \begin{tabular}{l c c c c}
    \hline
     & \multicolumn{2}{c}{\textbf{ClosedIE}} & \multicolumn{2}{c}{\textbf{OpenIE}} \\
    \textbf{LLM} & \textbf{Before} & \textbf{After} & \textbf{Before} & \textbf{After} \\
    \hline
    gpt-oss-120b & 91.5 & 95.9 & 25.3 & 68.6 \\
    Qwen3.8-27B  & 97.0 & 97.2 & 22.8 & 80.2 \\
    \hline
  \end{tabular}
  \caption{Effect of \textsc{Trallie}'s revised prompts on FDA 510(k) (100 documents, same LLMs and documents).
    \emph{Before}: original prompts; \emph{After}: schema discovery reuses the documents' own field labels at leaf level, and extraction copies values verbatim (capped at about 300 words) with null for missing fields.
    Gold field names in the discovered schema rose from 6/16 to 13/16 (gpt-oss-120b) and from 5/16 to 14/16 (Qwen3.8-27B).}
  \label{tab:prompt-ablation}
\end{table}

\begin{table*}[t]
  \centering
  \small
  \setlength{\tabcolsep}{4pt}
  \begin{tabular}{l c c c c c c c c c}
    \hline
     & & \multicolumn{4}{c}{\textbf{gpt-oss-120b}} & \multicolumn{4}{c}{\textbf{Qwen3.8-27B}} \\
     & & \textbf{ClosedIE} & \multicolumn{3}{c}{\textbf{OpenIE}} & \textbf{ClosedIE} & \multicolumn{3}{c}{\textbf{OpenIE}} \\
    \textbf{Website} & \textbf{\#Attr.} & \textbf{F1} & \textbf{P} & \textbf{R} & \textbf{F1} & \textbf{F1} & \textbf{P} & \textbf{R} & \textbf{F1} \\
    \hline
    \multicolumn{10}{l}{\textit{SWDE Movie}} \\
    AllMovie        & 59  & 94.0 & 88.0 & 60.1 & 71.5 & 94.0 & 90.2 & 65.7 & 76.0 \\
    AMC TV          & 19  & 97.5 & 66.9 & 70.8 & 68.8 & 97.5 & 76.9 & 80.2 & 78.6 \\
    Hollywood       & 278 & 99.2 & 42.8 & 40.7 & 41.7 & 99.6 & 50.7 & 46.8 & 48.7 \\
    iHeartMovies    & 12  & 89.5 & 46.2 & 39.2 & 42.4 & 89.6 & 46.9 & 47.2 & 47.1 \\
    IMDb            & 39  & 88.9 & 50.1 & 26.4 & 34.5 & 91.6 & 48.6 & 24.8 & 32.8 \\
    Metacritic      & 20  & 86.1 & 46.1 & 45.6 & 45.9 & 77.8 & 70.4 & 40.4 & 51.3 \\
    Rotten Tomatoes & 13  & 97.3 & 39.4 & 44.3 & 41.7 & 97.5 & 18.7 & 21.9 & 20.2 \\
    Yahoo Movies    & 13  & 93.5 & 49.4 & 55.4 & 52.2 & 99.5 & 63.4 & 70.2 & 66.6 \\
    \textit{Mean}   &     & 93.2 &      &      & 49.8 & 93.4 &      &      & 52.7 \\
    \hline
    \multicolumn{10}{l}{\textit{SWDE University}} \\
    CollegeProwler  & 23  & 96.6 & 41.5 & 45.1 & 43.2 & 97.0 & 56.9 & 60.7 & 58.7 \\
    eCampusTours    & 15  & 80.6 & 69.2 & 60.0 & 64.3 & 80.7 & 73.3 & 73.3 & 73.3 \\
    Embark          & 65  & 98.8 & 84.6 & 75.9 & 80.0 & 99.8 & 84.7 & 75.9 & 80.1 \\
    MatchCollege    & 66  & 82.7 & 47.0 & 19.6 & 27.7 & 85.9 & 65.0 & 26.7 & 37.8 \\
    US News         & 26  & 90.7 & 73.2 & 60.2 & 66.1 & 94.1 & 79.8 & 65.6 & 72.0 \\
    \textit{Mean}   &     & 89.9 &      &      & 56.3 & 91.5 &      &      & 64.4 \\
    \hline
  \end{tabular}
  \caption{Per-website SWDE results for \textsc{Trallie} on the first 100 pages of each website (no failed pages for either LLM).
    \#Attr.\ is the number of gold attributes, which is also $k$ for OpenIE; the discovered schema has 20 attributes, so on websites with more gold attributes the remainder cannot be matched.
    Means are over websites, as in Table~\ref{tab:main-results}.}
  \label{tab:swde-per-site}
\end{table*}

Table~\ref{tab:main-results} compares \textsc{Trallie} with the systems reported by \citet{Arora2023LanguageME} on the EVAPORATE benchmark. We report ClosedIE, where the attributes to extract are given, scored with token-level Text F1 \citep{rajpurkar2016squad}, and OpenIE, where the system must also discover the attributes, scored with Pair F1. We run \textsc{Trallie} with gpt-oss-120b \citep{openai2025gptoss} and Qwen3.8-27B \citep{qwen38}.



\textsc{Trallie} achieves the highest ClosedIE scores on every dataset with both LLMs
(Table~\ref{tab:main-results}), including on SWDE, which was held out during prompt
development. Unlike EVAPORATE-Direct \citep{Arora2023LanguageME}, which prompts once
per attribute, it extracts all attributes of a document in a single LLM call.
OpenIE is less uniform. \textsc{Trallie} outperforms EVAPORATE-Code+ on FDA 510(k) and,
with Qwen3.8-27B, on SWDE university websites, but trails it and DOM-LM
\citep{deng2022domlm} on movie websites. Per-website OpenIE scores range from 20.2 to
80.1 (Table~\ref{tab:swde-per-site}), and the analysis below shows that this gap comes
mainly from attribute naming rather than value extraction.

\paragraph{OpenIE Error Analysis.} Pair F1 counts a prediction as correct only if both the value and the attribute name match the gold annotation. Most of \textsc{Trallie}'s OpenIE errors on SWDE are naming errors rather than extraction errors: the discovered schema paraphrases a website's labels (e.g., \emph{director} where the gold attribute is \emph{Directed by}), so correctly extracted values are scored as wrong. To separate naming from extraction, we map each gold attribute to the predicted attribute whose values agree with it best and score the result with Text F1. Under this mapping, \textsc{Trallie} scores 78.0 (gpt-oss-120b) and 80.2 (Qwen3.8-27B) on movie websites, and 66.5 and 82.9 on university websites. This is a diagnostic only and is not comparable to the published Pair F1 numbers. The largest gap is on Rotten Tomatoes with Qwen3.8-27B: values are extracted well (92.1 under the mapping), but the schema uses names such as \emph{movie title} and \emph{release year} and includes near-duplicate attributes, giving a Pair F1 of 20.2. 

A second source of error is schema size: the discovered schema has 20 attributes, so on websites with more gold attributes (e.g., 278 on Hollywood and 66 on MatchCollege) the remainder cannot be matched. Because Pair recall is computed over (document, attribute, value) triples and many gold attributes are sparse, recall can still exceed the fraction of gold attributes covered by the schema. \citet{Arora2023LanguageME} also matched attribute names against a list of alternative valid names for SWDE that has not been released; our strict matching therefore likely understates \textsc{Trallie}'s OpenIE scores relative to the published ones.

\paragraph{Effect of prompt design.}
Table~\ref{tab:prompt-ablation} shows the effect of \textsc{Trallie}'s revised prompts on FDA 510(k), with the LLMs and documents held fixed. The largest effect is on OpenIE, where Pair F1 rises from 25.3 to 68.6
(gpt-oss-120b) and from 22.8 to 80.2 (Qwen3.8-27B). Most of this improvement comes from schema discovery: instructing the model to reuse the documents' own field labels at leaf level raises the number of gold attribute names recovered in the
discovered schema from 6 to 13 of 16 (gpt-oss-120b) and from 5 to 14 of 16
(Qwen3.8-27B). ClosedIE, where attribute names are given, changes less (91.5 to 95.9 and 97.0 to 97.2), mostly because extraction now copies values verbatim and returns null for missing fields. Because these prompts were developed on FDA 510(k), the FDA numbers may be optimistic; SWDE was held out during development, and the high ClosedIE scores there with both LLMs suggest the prompts are not specific to FDA 510(k).

\section{System Demonstration}

Our demonstration presents \textsc{Trallie} through an interactive Jupyter notebook and a command-line interface, and covers the following:

\begin{itemize}
\item Run live schema discovery and data extraction on sample document collections spanning scientific papers, regulatory filings, and web data
\item Compare extraction results across different LLM providers (e.g., GPT-4o, or local models such as gpt-oss, Qwen, and Llama) by changing a single provider argument
\item Observe the system's handling of long documents through overlapping chunking and LLM-based consolidation
\end{itemize}

After installation (i.e., \texttt{pip install trallie}), users can extract structured data from their own documents with a few lines of code. Sample documents and notebooks are available in the project repository,\footnote{\url{https://github.com/PiSchool/trallie}} and the live demo is available as a Colab notebook.\footnote{\url{https://colab.research.google.com/drive/124ECC8cQKD9zWSbPT9eB5Gcs-3UoB2vm}}

\section{Related Work}

Early open information extraction (OpenIE) systems extract relational tuples from text without a predefined schema, relying on self-supervision and linguistic structure \citep{banko2007open, angeli2015leveraging}. Many subsequent approaches for unstructured text use linguistic tools, such as dependency parsers \citep{Michael:25, tulpan-tsur:23}, to first induce structure and then apply heuristic rules on the resulting structure to extract information. Such pipelines perform poorly on long semi-structured documents such as PDFs \citep{Zhang:24}. Other methods assume the presence of human annotators in the loop to label the data and write the extraction rules \citep{schroeder2025justhumanloopinvestigating, Ding:25}, or assume access to domain-annotated training documents \citep{Kim:22, Meisenbacher:23}. For semi-structured web pages, the SWDE benchmark \citep{hao2011swde} has been used to study wrapper induction and zero-shot extraction from unseen websites \citep{lockard2020zeroshotceres}.

With large language models (LLMs), information extraction is increasingly formulated as a generation task, in which entities, relations, and attribute values are produced directly as text \citep{xu2023generativeie}. ChatIE \citep{wei2023chatie} shows that zero-shot extraction can be framed as a multi-turn question answering dialogue with a chat model. Closest to our setting, EVAPORATE \citep{Arora2023LanguageME} uses LLMs to generate structured tables from semi-structured document collections (e.g., HTML pages and PDFs) without domain-specific training or manual intervention. It identifies a schema by prompting the LLM on a small sample of documents and ranking the proposed attributes by frequency, and then populates the schema either by direct extraction, where the LLM processes each document individually, or by code synthesis, where the LLM writes extraction functions that are applied at scale. Its extension, EVAPORATE-Code+, combines multiple synthesized functions through weak supervision, which reduces computational cost.

Other work provides specialized models and libraries for schema-guided extraction. NuExtract \citep{nuextract24} is a family of lightweight models from NuMind, fine-tuned on synthetic data to fill a user-provided JSON template from the input, which can be used zero-shot or further fine-tuned; NuMind reports extraction performance comparable to much larger general-purpose models. LangExtract \citep{langextract2025} is an open-source Python library from Google that extracts structured information guided by a task description and a few examples, grounds each extracted value to its location in the source text, handles long documents through chunking and multiple extraction passes, and supports both cloud-hosted and local models. At the decoding level, constrained generation methods \citep{willard2023outlines} and provider-side JSON modes guarantee syntactically valid structured outputs, while leaving the definition of the schema to the user.

Finally, generalist named entity recognition models such as NuNER \citep{bogdanov2024nunerentityrecognitionencoder}, GLiNER \citep{zaratiana2023glinergeneralistmodelnamed}, UniversalNER \citep{zhou2024universalnertargeteddistillationlarge}, and GoLLIE \citep{sainz2024gollieannotationguidelinesimprove} extend extraction to entity types, or to annotation guidelines, not seen during training. These models operate on entity spans within a passage, whereas document-level attribute-value extraction also requires deciding which attributes describe a document as a whole.

\section{Conclusion}
We have introduced \textsc{Trallie}, a lightweight and modular open-source framework for transforming unstructured documents into structured, queryable formats. \textsc{Trallie} uses LLMs both to induce a schema from a document collection and to extract attribute values for that schema, reducing the need for manual schema design and domain-specific engineering. It processes PDF, HTML, TXT, and JSON documents and works with commercial, custom, and local LLM providers through a single configuration parameter. On the EVAPORATE benchmark, \textsc{Trallie} achieves the best ClosedIE scores on all three domains with both LLMs, and the best OpenIE scores on FDA 510(k) and, with Qwen3.8-27B, on SWDE university websites. On SWDE movie websites, OpenIE remains below EVAPORATE-Code+, mainly because discovered attribute names paraphrase the websites' labels, which suggests that further aligning schema names with the documents' own labels is the most promising direction for future work.

\section*{Limitations}
\paragraph{Comparability with EVAPORATE.}
EVAPORATE results are taken from \citet{Arora2023LanguageME}, since their LLM
(text-davinci-003) has been retired; differences therefore reflect both method and model.
We score \textsc{Trallie} with EVAPORATE's released evaluation code.
On SWDE, for cost reasons, we use the first 100 pages of each website rather than all
pages. Bootstrap intervals over pages (200 samples; \citep{efron1993bootstrap}) are
within $\pm$0.4 of each mean for both LLMs, but assume these pages are representative and ignore
variation from the choice of pages used for schema discovery.
EVAPORATE's list of alternative attribute names for SWDE is not public, so our strict
name matching may understate \textsc{Trallie}'s OpenIE scores.

\paragraph{Prompt development.}
\textsc{Trallie}'s prompts were revised on FDA 510(k), so those results may be
optimistic. SWDE was held out; there, ClosedIE gains persist with both LLMs, but OpenIE
exceeds EVAPORATE-Code+ \citep{Arora2023LanguageME} only on university websites with
Qwen3.8-27B and otherwise trails it, because discovered attribute names paraphrase
website labels.

\paragraph{Scope of the system.}
\textsc{Trallie} makes one LLM call per document, so unlike EVAPORATE-Code+, which
synthesizes extraction code once per collection, its cost grows linearly with
collection size. The discovered schema keeps only the most frequent attributes among a
small sample of documents (10 in our experiments), so long-tail attributes are missed
and schema quality depends on how representative that sample is. \textsc{Trallie}
extracts flat attribute--value records; it does not model relations, temporal
information or nested structure such as tables. Documents are converted to plain text
before extraction, which loses layout in multi-column PDFs and script-heavy HTML.
Outputs are checked only for well-formed JSON, not for value types or formats, so
high-stakes uses still require human review.

\section*{Acknowledgments}
We would like to thank all colleagues and collaborators who contributed to this work. This research was funded under the 4th Open Call of the NGI (Next Generation Internet) Search Initiative of the European Commission.

\bibliography{custom}

@article{Arora2023LanguageME,
  title   = {Language Models Enable Simple Systems for Generating Structured Views of Heterogeneous Data Lakes},
  author  = {Arora, Simran and Yang, Brandon and Eyuboglu, Sabri and Narayan, Avanika and Hojel, Andrew and Trummer, Immanuel and R{\'e}, Christopher},
  journal = {Proceedings of the VLDB Endowment},
  volume  = {17},
  number  = {2},
  pages   = {92--105},
  year    = {2023}
}

@inproceedings{hao2011swde,
  title     = {From One Tree to a Forest: A Unified Solution for Structured Web Data Extraction},
  author    = {Hao, Qiang and Cai, Rui and Pang, Yanwei and Zhang, Lei},
  booktitle = {Proceedings of the 34th International ACM SIGIR Conference on Research and Development in Information Retrieval},
  pages     = {775--784},
  year      = {2011}
}

@inproceedings{rajpurkar2016squad,
  title     = {{SQ}u{AD}: 100,000+ Questions for Machine Comprehension of Text},
  author    = {Rajpurkar, Pranav and Zhang, Jian and Lopyrev, Konstantin and Liang, Percy},
  booktitle = {Proceedings of the 2016 Conference on Empirical Methods in Natural Language Processing},
  pages     = {2383--2392},
  address   = {Austin, Texas},
  publisher = {Association for Computational Linguistics},
  year      = {2016},
  doi       = {10.18653/v1/D16-1264},
  url       = {https://aclanthology.org/D16-1264}
}

@article{deng2022domlm,
  title   = {{DOM-LM}: Learning Generalizable Representations for {HTML} Documents},
  author  = {Deng, Xiang and Shiralkar, Prashant and Lockard, Colin and Huang, Binxuan and Sun, Huan},
  journal = {arXiv preprint arXiv:2201.10608},
  year    = {2022}
}

@article{openai2025gptoss,
  title   = {gpt-oss-120b \& gpt-oss-20b Model Card},
  author  = {{OpenAI}},
  journal = {arXiv preprint arXiv:2508.10925},
  year    = {2025}
}

@misc{qwen38,
  title        = {{Qwen3.8-Max}: A New Bar for Coding and Cowork},
  author       = {{Qwen Team}},
  month        = aug,
  year         = {2026},
  howpublished = {\url{https://qwen.ai/blog?id=qwen3.8}},
  note         = {Model used: \url{https://huggingface.co/Qwen/Qwen3.8-27B}}
}

@book{efron1993bootstrap,
  title     = {An Introduction to the Bootstrap},
  author    = {Efron, Bradley and Tibshirani, Robert J.},
  publisher = {Chapman and Hall/CRC},
  address   = {New York},
  year      = {1993}
}

@inproceedings{banko2007open,
  title     = {Open Information Extraction from the Web},
  author    = {Banko, Michele and Cafarella, Michael J. and Soderland, Stephen and Broadhead, Matt and Etzioni, Oren},
  booktitle = {Proceedings of the 20th International Joint Conference on Artificial Intelligence (IJCAI)},
  pages     = {2670--2676},
  year      = {2007}
}

@inproceedings{angeli2015leveraging,
  title     = {Leveraging Linguistic Structure for Open Domain Information Extraction},
  author    = {Angeli, Gabor and Johnson Premkumar, Melvin Jose and Manning, Christopher D.},
  booktitle = {Proceedings of the 53rd Annual Meeting of the Association for Computational Linguistics (ACL)},
  pages     = {344--354},
  year      = {2015}
}

@article{Michael:25,
  title   = {A Fundamental Algorithm for Dependency Parsing (with Corrections)},
  author  = {Covington, Michael A.},
  journal = {arXiv preprint arXiv:2510.19996},
  year    = {2025},
  url     = {https://arxiv.org/abs/2510.19996}
}

@inproceedings{tulpan-tsur:23,
  title     = {A Deeper (Autoregressive) Approach to Non-Convergent Discourse Parsing},
  author    = {Tulpan, Yoav and Tsur, Oren},
  booktitle = {Proceedings of the 2023 Conference on Empirical Methods in Natural Language Processing},
  pages     = {12883--12895},
  publisher = {Association for Computational Linguistics},
  year      = {2023}
}

@article{Zhang:24,
  title   = {Document Parsing Unveiled: Techniques, Challenges, and Prospects for Structured Information Extraction},
  author  = {Zhang, Qintong and Wang, Bin and Huang, Victor Shea-Jay and Zhang, Junyuan and Wang, Zhengren and Liang, Hao and He, Conghui and Zhang, Wentao},
  journal = {arXiv preprint arXiv:2410.21169},
  year    = {2024},
  url     = {https://arxiv.org/abs/2410.21169}
}

@article{schroeder2025justhumanloopinvestigating,
  title   = {Just Put a Human in the Loop? Investigating {LLM}-Assisted Annotation for Subjective Tasks},
  author  = {Schroeder, Hope and Roy, Deb and Kabbara, Jad},
  journal = {arXiv preprint arXiv:2507.15821},
  year    = {2025},
  url     = {https://arxiv.org/abs/2507.15821}
}

@article{Ding:25,
  title   = {Computational Review of Technology-Assisted Medical Evidence Synthesis through Human-{LLM} Collaboration: A Case Study of {Cochrane}},
  author  = {Ding, Yiping and Jiang, Xiaorui and Akinseloyin, Opeoluwa},
  journal = {medRxiv preprint 2025.11.08.25339805},
  year    = {2025},
  doi     = {10.1101/2025.11.08.25339805}
}

@article{Kim:22,
  title   = {{OASYS}: Domain-Agnostic Automated System for Constructing Knowledge Base from Unstructured Text},
  author  = {Kim, Minsang and Je, Sang-hyun and Park, Eunjoo},
  journal = {arXiv preprint arXiv:2207.07597},
  year    = {2022},
  url     = {https://arxiv.org/abs/2207.07597}
}

@article{Meisenbacher:23,
  title   = {Transforming Unstructured Text into Data with Context Rule Assisted Machine Learning ({CRAML})},
  author  = {Meisenbacher, Stephen and Norlander, Peter},
  journal = {arXiv preprint arXiv:2301.08549},
  year    = {2023},
  url     = {https://arxiv.org/abs/2301.08549}
}

@inproceedings{lockard2020zeroshotceres,
  title     = {{ZeroShotCeres}: Zero-Shot Relation Extraction from Semi-Structured Webpages},
  author    = {Lockard, Colin and Shiralkar, Prashant and Dong, Xin Luna and Hajishirzi, Hannaneh},
  booktitle = {Proceedings of the 58th Annual Meeting of the Association for Computational Linguistics (ACL)},
  year      = {2020}
}

@article{xu2023generativeie,
  title   = {Large Language Models for Generative Information Extraction: A Survey},
  author  = {Xu, Derong and Chen, Wei and Peng, Wenjun and Zhang, Chao and Xu, Tong and Zhao, Xiangyu and Wu, Xian and Zheng, Yefeng and Wang, Yang and Chen, Enhong},
  journal = {arXiv preprint arXiv:2312.17617},
  year    = {2023}
}

@article{wei2023chatie,
  title   = {Zero-Shot Information Extraction via Chatting with {ChatGPT}},
  author  = {Wei, Xiang and Cui, Xingyu and Cheng, Ning and Wang, Xiaobin and Zhang, Xin and Huang, Shen and Xie, Pengjun and Xu, Jinan and Chen, Yufeng and Zhang, Meishan and Jiang, Yong and Han, Wenjuan},
  journal = {arXiv preprint arXiv:2302.10205},
  year    = {2023}
}

@misc{nuextract24,
  title        = {{NuExtract} 1.5: Multilingual, Infinite Context, Still Small, and Better than {GPT-4o}!},
  author       = {{NuMind}},
  year         = {2024},
  howpublished = {\url{https://www.numind.ai/blog/nuextract-1-5---multilingual-infinite-context-still-small-and-better-than-gpt-4o}}
}

@misc{langextract2025,
  title        = {{LangExtract}: A Python Library for Extracting Structured Information from Unstructured Text Using {LLMs}},
  author       = {{Google}},
  year         = {2025},
  howpublished = {\url{https://github.com/google/langextract}}
}

@article{willard2023outlines,
  title   = {Efficient Guided Generation for Large Language Models},
  author  = {Willard, Brandon T. and Louf, R{\'e}mi},
  journal = {arXiv preprint arXiv:2307.09702},
  year    = {2023}
}

@inproceedings{bogdanov2024nunerentityrecognitionencoder,
  title     = {{NuNER}: Entity Recognition Encoder Pre-training via {LLM}-Annotated Data},
  author    = {Bogdanov, Sergei and Constantin, Alexandre and Bernard, Timoth{\'e}e and Crabb{\'e}, Benoit and Bernard, Etienne},
  booktitle = {Proceedings of the 2024 Conference on Empirical Methods in Natural Language Processing},
  publisher = {Association for Computational Linguistics},
  year      = {2024}
}

@inproceedings{zaratiana2023glinergeneralistmodelnamed,
  title     = {{GLiNER}: Generalist Model for Named Entity Recognition using Bidirectional Transformer},
  author    = {Zaratiana, Urchade and Tomeh, Nadi and Holat, Pierre and Charnois, Thierry},
  booktitle = {Proceedings of the 2024 Conference of the North American Chapter of the Association for Computational Linguistics: Human Language Technologies (NAACL)},
  publisher = {Association for Computational Linguistics},
  year      = {2024}
}

@inproceedings{zhou2024universalnertargeteddistillationlarge,
  title     = {{UniversalNER}: Targeted Distillation from Large Language Models for Open Named Entity Recognition},
  author    = {Zhou, Wenxuan and Zhang, Sheng and Gu, Yu and Chen, Muhao and Poon, Hoifung},
  booktitle = {The Twelfth International Conference on Learning Representations (ICLR)},
  year      = {2024}
}

@inproceedings{sainz2024gollieannotationguidelinesimprove,
  title     = {{GoLLIE}: Annotation Guidelines Improve Zero-Shot Information-Extraction},
  author    = {Sainz, Oscar and Garc{\'i}a-Ferrero, Iker and Agerri, Rodrigo and Lopez de Lacalle, Oier and Rigau, German and Agirre, Eneko},
  booktitle = {The Twelfth International Conference on Learning Representations (ICLR)},
  year      = {2024}
}




\end{document}